\pdfoutput=1
\documentclass{article}
\usepackage{ijcai16}
\usepackage{multirow}
\usepackage{times}
\usepackage{amsmath}

\DeclareMathOperator*{\argmax}{arg\,max}

\title{Exploiting N-Best Hypotheses to Improve an SMT Approach to Grammatical Error Correction}

\author{
Duc Tam Hoang$^{1}$ \and Shamil Chollampatt$^{2}$ \and Hwee Tou Ng$^{1,2}$ \\
$^{1}$Department of Computer Science \\
$^{2}$NUS Graduate School for Integrative Sciences and Engineering \\
National University of Singapore \\
\{hoangdt, nght\}@comp.nus.edu.sg, shamil@u.nus.edu
}

\def \citep #1{\citeauthor{#1}~\shortcite{#1}}

\begin{document}
    \maketitle
    
    \begin{abstract}
        Grammatical error correction (GEC) is the task of detecting and correcting grammatical errors in texts written by second language learners. The statistical machine translation (SMT) approach to GEC, in which sentences written by second language learners are translated to grammatically correct sentences, has achieved state-of-the-art accuracy. However, the SMT approach is unable to utilize global context. In this paper, we propose a novel approach to improve the accuracy of GEC, by exploiting the n-best hypotheses generated by an SMT approach. Specifically, we build a classifier to score the edits in the n-best hypotheses. The classifier can be used to select appropriate edits or re-rank the n-best hypotheses. We apply these methods to a state-of-the-art GEC system that uses the SMT approach. Our experiments show that our methods achieve statistically significant improvements in accuracy over the best published results on a benchmark test dataset on GEC. 
    \end{abstract}

    \section{Introduction}
    \label{sec:intro}
    
    
    Using statistical machine translation (SMT) for the task of grammatical error correction (GEC) has gained popularity as one of the most promising approaches. In general, language learners can make different types of grammatical errors. It becomes difficult to target each error type with traditional classification-based approaches that use hand-crafted features for correcting specific error types. On the other hand, SMT based systems have the ability to correct a wide variety of error types without language-specific features. In order to use the SMT framework, the task of GEC is formulated as a translation task from \emph{bad} English to \emph{good} English. The SMT system, which is trained using parallel corpora (the erroneous sentences and their corresponding corrected sentences), attempts to output the corrected target sentence given an erroneous source sentence. 
    
    Phrase-based SMT components have been used in state-of-the-art grammatical error correction systems \cite{Susanto2010}. It has been shown that increasing the amount of parallel data further improves the performance of SMT-based GEC systems \cite{grundkiewicz2014amu}. However, SMT systems excessively rely on parallel data and tend to correct errors depending on the frequencies of corrections observed in the training data, without considering a global context. The error correction can be further improved if it had access to a larger context spanning phrase boundaries. Specifically, the phrase-based SMT approach does not capture long-distance dependency between words that are further apart in a sentence. As a consequence, the system will make several independent corrections, yet produce an overall ungrammatical sentence. 
    
    Table~\ref{tab:errorexample} shows a few examples of invalid corrections made by the SMT approach. In the first example, \textit{friends feels lonely} is corrected to \textit{friends feel lonely}. Although this correction seems accurate locally, it is actually invalid considering the global context in the sentence. Specifically, the SMT approach has limitations in correcting grammatical errors like subject-verb agreement, article-noun, and verb-form errors.  The examples in Table~\ref{tab:errorexample} show that taking the global context into consideration can improve the SMT approach to GEC. 
    
    \begin{table*}[ht!]
        \begin{center}
            \begin{tabular}{ | l | l | l | }
                \hline
                \# & {\bf Description} & {\bf Sentence}\\ \hline\hline 
                1 & \textit{input} & The man with a huge number of friends feels lonely .\\
                & \textit{reference} & The man with a huge number of friends feels lonely .\\  
                & \textit{hypothesis} & The man with a huge number of friends \underline{feel} lonely . \\ \hline
                2 & \textit{input} & He carries a gun into his pocket and walk into the bar .\\ 
                & \textit{reference} & He carries a gun \underline{in} his pocket and \underline{walks} into the bar .\\ 
                & \textit{hypothesis} & He carries a gun \underline{in} his pocket and \underline{walking} into the bar .\\ \hline
                3 & \textit{input} & He has an carved green Venetian glass salad bowls .\\
                & \textit{reference} & He has \underline{\hspace{0.4cm}} carved green Venetian glass salad bowls .\\
                & \textit{hypothesis} & He has \underline{a} carved green Venetian glass salad bowls .\\ \hline
                4 & \textit{input} & The train crashed and all passengers were died . \\
                & \textit{reference} & The train crashed and all passengers \underline{\hspace{0.4cm}} died . \\				
                & \textit{hypothesis} & The train \underline{crashes} and all passengers \underline{\hspace{0.4cm}} died . \\ \hline			
            \end{tabular} \par
            \caption[Table]{Examples of errors made by an SMT approach to GEC.}
            \label{tab:errorexample}
        \end{center}
    \end{table*}
    
    In this paper, we improve the correction made by an SMT approach to GEC, by incorporating a larger context and adding linguistic information. Our {\it novel} approach exploits the top $n$ correction candidates (\textit{n-best} hypotheses) generated by the SMT system. We build an \textit{edit classifier} to validate the edits made by the SMT system. The features of this classifier use the available global context from a sentence. We propose two different ways of using this classifier to improve the corrections made by the SMT system:
    \begin{enumerate}
        \item \textit{\textbf{Reranking}}: We rerank the n-best hypotheses by assigning a score to each hypothesis. The score is used as a feature along with other SMT features in order to rerank the n-best hypotheses.
        \item \textit{\textbf{Edit selection}}: All the edits in the n-best hypotheses are scored by the classifier. We then select the highest scoring edits from different hypotheses and apply them on the source sentence in order to generate a better hypothesis.  
        
    \end{enumerate}

    The remainder of this paper is organized as follows. Section~\ref{sec:rel} discusses the related work. Section~\ref{sec:method} describes the \textit{edit classifier}.   Section~\ref{sec:exploit} describes the two methods to exploit the n-best hypotheses. Section~\ref{sec:experiment} presents the experimental setup and results. Section~\ref{sub:discuss} discusses the results. Finally, Section~\ref{sec:con} gives the conclusion.

    \section{Related Work}
    \label{sec:rel}
    
    Earlier approaches to GEC aimed at building classifiers or rule-based systems targeting specific error types. More recently, the SMT approach to GEC has gained attention.  \citep{brockett2006correcting} used an SMT approach to correct errors related to countability of mass nouns, which poses problems to English as Second Language (ESL) learners.  \citep{mizumoto2011mining} used an SMT approach that focuses on correcting grammatical errors of Japanese learners. They used large-scale data mined from the language learning website, Lang-8\footnote{http://lang-8.com} in order to train their SMT system. They later implemented an SMT system for GEC in English ~\cite{tajiri2012tense}. The availability of large-scale error annotated data for GEC further increased the popularity of the SMT approach to GEC. \citep{grundkiewicz2014amu} performed various experiments using an SMT approach to GEC, such as increasing the size of the parallel corpora, using web-scale language models, and varying the methods for tuning the SMT system.
    
    Several shared tasks have been organized in recent years for the GEC task. The CoNLL-2013 shared task~\cite{Ng2013} focuses on correction of five error types: determiner, preposition, noun number, verb form, and subject-verb agreement errors. \citep{yuan2013constrained} added other learner corpora to the provided training data. They used a phrase-based SMT system for correction of all five error types. Meanwhile, \citep{yoshimoto2013naist} used an SMT approach for only two error types, prepositions and determiners. For other error types, they reported that the classification approach performs better. The CoNLL-2014 shared task~\cite{Ng2014} focused on correction of all error types. The number of teams using the SMT approach increased. They achieved competitive results compared to other teams which used the classification approach~\cite{rozovskaya2014illinois}. 
    
    Recent work by \citep{Susanto2010} combined the outputs of multiple systems based on both the SMT approach and the classification approach. They used MEMT~\cite{heafield2010combining} to combine four systems (two classification-based systems and two SMT-based systems) to achieve an $F_{0.5}$ score of 39.39.  \citep{Zheng2016neural} later proposed a neural machine translation approach for grammatical error correction. They achieved a higher $F_{0.5}$ score of 39.90 on the CoNLL-2014 test set, but using a much larger training set. 
    
	More recently, discriminative reranking methods for SMT-based GEC have been proposed \cite{Mizumoto2016,Zheng2016reranking}. \citep{Zheng2016reranking} use a ranking SVM to rerank the n-best list without employing any syntactic features. Syntactic features and long-distance dependency relationships are important cues to assess the grammaticality of a sentence. Hence, we make use of syntactic features like constituency parse and part-of-speech tags in our approach. \citep{Mizumoto2016} also perform discriminative reranking with a perceptron using syntactic features. However, they only include the final score given by the SMT system in reranking and do not include all the other SMT features.  On the other hand, we perform reranking using all SMT features and an additional feature given by our edit classifier. Most importantly, both \citep{Mizumoto2016} and \citep{Zheng2016reranking} focus solely on reranking the n-best hypotheses and do not generate any new hypothesis, unlike our \textit{edit selection} method. 
    \section{Edit Classifier}
    \label{sec:method}
    
    In general, a grammatical error correction system takes an input \textit{source} sentence and produces a corrected \textit{hypothesis} sentence. We extract the corrections made by the system, which we refer to as \textit{edits}. For example: \\
    \textit{Source}: \\
    \hspace{0.5cm} He carries a gun \underline{into} his pocket and \underline{walk} into the bar.\\
    \textit{Hypothesis}:\\
    \hspace{0.5cm} He carries a gun \underline{in} his pocket and \underline{walking} into the bar.\newline
    
    The \textit{edits} in the above example are ($into \rightarrow in$) and ($walk \rightarrow walking$). We build a binary classifier to classify an \textit{edit} as valid or invalid.  A valid edit is a good correction and an invalid edit is a bad correction made by the system. In this example, the first edit is valid while the second is not. Section \ref{sub:features} describes the features used to train the classifier and Section \ref{sub:cw} describes the training algorithm.

    \subsection{Features}
    \label{sub:features}
    We select features to build a binary classifier to distinguish between valid and invalid edits.  We use both categorical and numerical features. 
    If the same edit occurs in more than one hypothesis, the features are extracted from the highest ranked hypothesis in the n-best list.
    Table~\ref{tab:features2} shows the detailed features using an illustrating example. We use a 5-gram language model trained on English Wikipedia (approximately 1.78 billion words). We use the Stanford parser~\cite{klein2003} for constituency parsing. 
    
    \begin{table}[ht!]
        \small
        \begin{center}
            \begin{tabular}{ | l | l | }
                \hline
                {\bf Features} & {\bf Example}\\ \hline\hline 
                \multicolumn{2}{|l|}{\it SMT Features}  \\ \hline
                rank of the hypothesis & $1$ \\ \hline
                
                \multicolumn{2}{|l|}{\it Lexical and POS Features}  \\ \hline
                source phrase & $waits$ \\
                hypothesis phrase & $sat$ \\
                source + hyp. phrase & $waits$+$sat$\\
                POS of source phrase & $VBZ$ \\
                POS of hyp. phrase & $VBD$ \\
                POS of source + hyp. phrase & $VBZ$+$VBD$ \\ \hline
                
                \multicolumn{2}{|l|}{\it Context Features}  \\ \hline
                word before + source phrase & $cat$+$waits$ \\
                word before + hyp. phrase & $cat$+$sat$ \\
                source phrase + word after & $waits$+$on$ \\
                hyp. phrase + word after & $sat$+$on$ \\
                
                nearest NP head to the left & $cat$  \\
                {\hspace{0.2cm}} + (source, hyp. phrase) & $cat$+$\{waits,sat\}$ \\ 
                nearest NP head to the right & $dog$ \\
                {\hspace{0.2cm}} + (source, hyp. phrase) & $dog$+$\{waits,sat\}$ \\ 
                nearest VP head to the left & $NULL$  \\
                {\hspace{0.2cm}} + (source, hyp. phrase) & $NULL$+$\{waits,sat\}$ \\                
                nearest VP head to the right & $eats$ \\ 
                {\hspace{0.2cm}} + (source, hyp. phrase) & $eats$+$\{waits,sat\}$ \\ \hline
                \multicolumn{2}{|l|}{\it Language Model Features}  \\ \hline
                source sentence & $log\ LM(src)$\\
                hypothesis sentence & $ log\ LM(hyp)$\\
                source phrase & $ log\ LM(waits)$\\
                hypothesis phrase & $log\ LM(sat)$\\
                word before+source phrase & $ log\ LM(cat\ waits)$\\
                word before+hyp. phrase & $log\ LM(cat\ sat)$\\
                source phrase+word after & $ log\ LM(waits\ on)$\\
                hyp. phrase + word after & $log\ LM(sat\ on)$\\
                \multirow{2}{*}{score difference} & $log\ LM(sat) - log LM(waits) $ \\
                & $log\ LM(hyp) - log LM(src)$ \\ \hline
                
            \end{tabular} \par
            \caption[Table]{ Features for the edit classifier. Example: edit $waits \rightarrow sat$ from the edits \textit{``the cat \{\underline{waits}$\rightarrow$\underline{sat}\} on the dog and eats a mouse .''} }
            \label{tab:features2}
        \end{center}
    \end{table}
    
    There are four main categories of features which are used to build the model, described as follows:
    
    \subsubsection*{SMT Features}
    
    We use the rank of the hypothesis in the n-best list in which the edit occurs as a feature.  Our intuition is that a hypothesis with a higher rank may contain more valid edits than one with a lower rank.
    
    \subsubsection*{Lexical and POS Features}
    
    We use the words occurring in the source side and target side of an edit and their parts-of-speech (POS) as features. The lexical features can determine the choice and order of words and the POS features can determine the grammatical roles of words in the edit within a hypothesis.
    
    \subsubsection*{Context Features}
    
    Context features correspond to how the edits interact with other words of the sentence. They capture the context information surrounding the current edit, ranging from neighboring words (preceding and following words) to constituents (heads of noun phrases and verb phrases). The neighboring words provide collocation information. We identify the interaction between the edit and constituents by the nearest noun phrase (or verb phrase) head to the left (or right) of the edit. For example: between verb and preposition (\textit{\underline{send} a huge package \underline{to} my friend}), between subject and verb (\textit{\underline{people} with albinism \underline{are} prone to sunburn)}, and between verb forms (\textit{He \underline{opened} the windows and she \underline{felt} cold}).
    
    \subsubsection*{Language Model Features}
    Additionally, we include the language model scores for the source and target phrases of the \textit{edit} and the source sentence and hypothesis sentence after applying the \textit{edit}. The language model features will help to ensure fluency of the hypothesis sentence after applying the \textit{edit}.

    \subsection{Training the Classifier}
    \label{sub:cw}
    
    In order to obtain edits to train the classifier, we use an SMT-based GEC system to translate the source side of an ESL corpus and obtain the n-best hypotheses for each source sentence. We use the NUS Corpus of Learner English (NUCLE) \cite{dahlmeier2013building} as the ESL corpus.  We obtain the edits of each hypothesis for every source sentence.  The edits which match the gold edits are considered valid and the other edits are considered invalid. Gold edits are obtained by comparing a corrected sentence by human to the original source sentence written by an ESL learner. 
    
    We choose confidence-weighted learning~\cite{dredze2008confidence} to train the \textit{edit classifier}. Confidence-weighted (CW) learning is an online learning algorithm which outputs a real number corresponding to the confidence in the prediction. CW weighted classifiers perform well in cases where the feature space is high dimensional and sparse, making it a popular choice for natural language processing tasks which use n-grams as features. Furthermore, to improve our classifier, we tuned a threshold for our classifier using the CoNLL-2013 dataset and performed a grid search for values of threshold from --0.5 to 0.5 with a step size of 0.01.

    \section{Exploiting N-Best Hypotheses}
    \label{sec:exploit}
    
    We use a phrase-based SMT approach for GEC. It follows the log-linear model formulation~\cite{och2002discriminative}: 
    
    \begin{eqnarray}
    \hat{t} = \argmax_{t}\ \exp\Big(\sum_{i=1}^{n} \lambda_{i} h_{i}(s,t)\Big)
    \label{eq:smt-formula}
    \end{eqnarray}
    
    In Equation~\ref{eq:smt-formula}, $s$ is the source sentence, $t$ is a hypothesis sentence, $n$ is the number of features, $h_{i}$ is a feature function, and $\lambda_{i}$ is its weight. The feature weights $\lambda_i$ is tuned by minimum error rate training (MERT) ~\cite{och2003minimum} on a development set optimizing the $F_{0.5}$ measure \cite{grundkiewicz2014amu}. The $F_{0.5}$ measure is a popular metric used for GEC \cite{dahlmeier2012better} and has been used in the CoNLL-2014 shared task as the official evaluation metric.
    
    The SMT decoder finds the best hypothesis $\hat{t}$ according to Equation~\ref{eq:smt-formula}. Alternatively, the SMT system can produce a list of n-best hypotheses ranked by the log-linear model. We propose two methods of using these n-best hypotheses to produce a better corrected output: \textit{reranking} and \textit{selection of edits}. Reranking is described in Section \ref{sec:reranking} and edit selection in Section \ref{sec:edit-selection}
    
    \subsection{Reranking}
    \label{sec:reranking}
    A potential use of the edit classifier is to rerank the SMT n-best hypotheses. We obtain the n-best hypotheses using an SMT system. We augment the list of features for each hypothesis with the average score given by the edit classifier for all edits in the hypothesis. The new feature function $h_{n+1}$ is given by the following equation:
    
    \begin{eqnarray}
    h_{n+1}(s,t) = \frac{1}{|E|} \sum_{e \in E} score(e)
    \end{eqnarray}
    where $s$ and $t$ are the source sentence and the hypothesis sentence respectively, $E$ is the set of edits associated with hypothesis $t$, and  $score(e)$ is the score assigned to the edit $e$ by the edit classifier. The weight $\lambda_{n+1}$ for this feature function is optimized along with the other SMT features by running MERT optimizing $F_{0.5}$ on the development set. We then rescore and rerank the hypotheses with the new set of features and weights.

    \subsection{Edit Selection}
    \label{sec:edit-selection}
    
    The edit classifier is used to classify an edit as valid or invalid. We discard all invalid edits. We adopt a simple scheme to select a subset of valid edits. From the set of all valid edits of all n-best hypotheses, we first select the edit which has the highest score assigned by the edit classifier. We then select the next highest scoring edit from the set, on the condition that it does not overlap with any edits selected so far. Two edits overlap if the source phrases of the two edits have some overlapping words in the source sentence. We repeat this process until we cannot find any additional non-overlapping edit. Compared to the reranking method which is restricted to reranking the top n hypotheses but cannot generate a new hypothesis sentence, the edit selection method is capable of generating a new hypothesis sentence by selecting edits from different hypotheses.

    \section{Experiments}
    \label{sec:experiment}
    
    We empirically evaluate our proposed methods in the context of the CoNLL-2014 shared task. 
    Description of the shared task can be found in the overview paper~\cite{Ng2014}. In this section, we only describe the salient details related to our work.
    
    \subsection{Setup}
    \label{sec:setup}
    
    We build a baseline system using Moses~\cite{koehn2007moses} with the standard configuration for phrase-based SMT. Two parallel corpora are used for training the translation model. The first corpus is the NUS Corpus of Learner English (NUCLE) ~\cite{dahlmeier2013building}. The second corpus is the Lang-8 Corpus of Learner English v1.0~\cite{tajiri2012tense}. We train one phrase table using the concatenation of the two corpora. Word alignments are obtained using GIZA++~\cite{och2003systematic} and phrase pairs are extracted using the standard phrase extraction tool provided by Moses. Statistics of the training data can be found in Table~\ref{tab:train-data}.
    
    We train two 5-gram language models using the KenLM language modeling toolkit~\cite{Heafield-kenlm}. The first language model is trained on the target side of NUCLE. The second language model is trained on English Wikipedia (approximately 1.78 billion words). The two language models are used as two separate features in the log-linear model.
    
    We do not use any reordering model as most error types do not involve long-distance reordering. Local reordering is generally captured by phrase pairs in the phrase table. The distortion limit is set to $0$ to prevent reordering during the hypothesis generation phase. Tuning is done on the CoNLL-2013 data. We use MERT~\cite{och2003minimum} as the tuning method.
    
    \begin{table}[!ht]
        \centering
        \begin{tabular}{|l|r|r|r|}
            \hline
            \textbf{Dataset} & \textbf{\# sentences} & \textbf{\begin{tabular}[c]{@{}l@{}}\# source\\ tokens\end{tabular}} & \textbf{\begin{tabular}[c]{@{}l@{}}\# target\\ tokens\end{tabular}}         \\ \hline
            \hline
            NUCLE            & 57,151           & 1,161,567                & 1,155,559         \\
            \hline
            Lang-8 v1.0      & 1,114,139        & 12,945,666               & 13,232,058        \\
            \hline
            CoNLL-2013       & 1,381            & 29,207                   & 28,743            \\
            \hline
            CoNLL-2014 & 1,312 & 30,114 & \begin{tabular}[c]{@{}l@{}}29,881\\ 30,229\end{tabular} \\
            \hline
        \end{tabular}
        \caption{Statistics of training, development, and test data. The CoNLL-2014 test data has corrected sentences by two human annotators, hence it has two target token counts, one for the corrections made by each annotator.}
        \label{tab:train-data}
    \end{table}       
    Finally, the system is evaluated on the CoNLL-2014 test set. Statistics of the development and test data can be found in Table~\ref{tab:train-data}, in which the CoNLL-2014 test data has corrected sentences by two human annotators. We use the $F_{0.5}$ measure as the evaluation metric for both tuning and testing. $F_{0.5}$ weights precision twice as much as recall. Given a set of sentences, where $G_s$ is the set of gold edits in the annotation for sentence $s$, and $E_s$ is the set of system edits for sentence s, precision, recall and $F_{0.5}$ are defined as follows:
    \begin{equation}
    P = \frac{ \sum_{s} |G_s \cap E_s| }{\sum_{s}|E_s|}
    \end{equation}    
    \begin{equation}
    R = \frac{ \sum_{s} |G_s \cap E_s| }{\sum_{s}|G_s|}
    \end{equation}
    
    and
    \begin{equation}
    F_{0.5} = \frac{  ( 1  + 0.5^2) \times R \times P }{ R + 0.5^2 \times P}
    \end{equation}
    where $G_s \cap E_s$ is defined as:
    \begin{equation}
    G_s \cap E_s = \{e \in E_s \ | \ \exists g \in G_s, match(g,e) \}
    \end{equation}
    We use the M$^2$Scorer~\cite{dahlmeier2012better} for evaluation, which was the official scorer for the CoNLL-2014 shared task. The scorer determines the system edits that have maximal overlap with the gold edits. For the statistical significance test, we use one-tailed sign test with bootstrap resampling on 100 samples.

    \subsection{Results}
    
    We first evaluate our edit classifier separately, independent of the entire GEC system. In order to do this, we compute the accuracy of the classifier on the edits obtained from the CoNLL-2013 dataset. Since we use the CoNLL-2013 dataset for threshold tuning, we evaluate the classifier before threshold tuning. We perform ablation testing, by removing each category of features: SMT features, lexical and POS features, context features, and language model features. The results of our experiments are summarized in Table \ref{tab:feature-selection}, showing that each group of features contributes to the performance of our classifier.

    \begin{table}[!h]
        \centering
        \begin{tabular}{|l|c|c|c|}
            \hline
            {\bf Features} & {\bf Accuracy} \\ \hline
            All	& {\bf 61.25} \\ 
            without SMT features	         & 61.17 \\
            without lexical and POS features & 61.01 \\
            without context features &	60.92 \\
            without LM features   &	59.93 \\ \hline
        \end{tabular}
        \caption{Feature ablation test. }
        \label{tab:feature-selection}
    \end{table}

    Table~\ref{tab:baseline} shows the performance of the baseline SMT system on the CoNLL-2014 test set. An important point is that the baseline system outperforms all state-of-the-art systems on the CoNLL-2014 test set, making it a highly competitive baseline. Note that our baseline system uses exactly the same training data as \cite{Susanto2010} for training the translation model and the language model. The difference between our baseline system and the SMT components of \cite{Susanto2010} is that we tune with $F_{0.5}$ instead of BLEU and we use the standard Moses configuration without the Levenshtein distance feature.
    
    \begin{table}[!ht]
        \begin{center}
            \begin{tabular}{| l | c | c | c |}
                \hline
                {\bf System} & {\bf P} & {\bf R}  & {\bf F$_{0.5}$} \\ \hline\hline
                Baseline & 50.56 & 22.68 & {\bf40.58} \\ 
                \hline
                \multicolumn{4}{|l|}{\bf State-of-the-art systems} \\\hline
                \small{\citep{Mizumoto2016}} & 45.80 & 26.60 & 40.00 \\ \hline
                 \small{\citep{Zheng2016neural}} & - & - & 39.90 \\ \hline
                 \small{\citep{Susanto2010}} & 53.55 & 19.14 & 39.39 \\ \hline
                 \small{\citep{Zheng2016reranking}} & - & - & 38.08 \\ \hline
                \multicolumn{4}{|l|}{\bf Top 3 systems in the CoNLL-2014 shared task} \\\hline
                CAMB & 39.71 & 30.10 & 37.33 \\ 
                CUUI & 41.78 & 24.88 & 36.79 \\
                AMU & 41.62 & 21.40 & 35.01 \\\hline
            \end{tabular}
            \caption{Performance of the baseline system and other state-of-the-art systems on the CoNLL-2014 test set.}
            \label{tab:baseline}
        \end{center}
    \end{table}

    We perform reranking and edit selection using the edit classifier. The experimental results on the n-best hypotheses generated by the SMT baseline on the CoNLL-2014 test set are shown in Table~\ref{tab:post-processing}. The reranking method is applied on the 5-best and 10-best hypotheses. The edit selection method selects non-overlapping edits from the top $n$ hypotheses with $n \in [1,5]$.

    \begin{table}[!ht]
        \centering
        \begin{tabular}{|l|c|c|l|}
            \hline
            {\bf System} & \textbf{P} & \multicolumn{1}{c|}{\textbf{R}} & \multicolumn{1}{c|}{\textbf{F$_{0.5}$}} \\ \hline
            \multicolumn{1}{|l|}{Baseline}    & 50.56 & 22.68 & 40.58	\\ \hline
            \multicolumn{4}{|l|}{\textbf{Reranking}} \\ \hline
            \multicolumn{1}{|l|}{\textit{5-best}}    & 50.32	& 22.99 & 40.65	\\ \hline            
            \multicolumn{1}{|l|}{\textit{10-best}}    & 50.79	& 22.92 & {\bf 40.85\textsuperscript{*}}	\\ \hline
            \multicolumn{4}{|l|}{\textbf{Edit selection}} \\ \hline
            \multicolumn{1}{|l|}{\textit{1-best}}    & 51.22	& 22.28 &	40.66   \\
            \multicolumn{1}{|l|}{\textit{2-best}}    & 50.35 & 23.70 & 41.11  \\
            \multicolumn{1}{|l|}{\textit{3-best}}    &  50.31 & 23.82 &	41.16  \\
            \multicolumn{1}{|l|}{\textit{4-best}}    &  50.31 & 23.82 &	41.16  \\
            \multicolumn{1}{|l|}{\textit{5-best}}    & 50.35 &	23.84	& {\bf 41.19\textsuperscript{*}}	 \\	\hline
        \end{tabular}
        \caption{Performance of reranking and edit selection on the CoNLL-2014 test set. \textsuperscript{*} indicates that the improvements are statistically significant ($p < 0.01$) compared to the baseline system.}
        \label{tab:post-processing}
    \end{table}
    
    \begin{table*}[ht!]
        \small
        \begin{center}
            \begin{tabular}{ |p{2cm}|p{14cm}| }
                \hline
                {\bf Description} & {\bf Sentence}\\ \hline\hline 

                \multicolumn{2}{|l|}{\textit{Subject verb agreement}} \\ \hline
            \textit{source} & Unlike people with high proportion of skin pigment , {\bf people} with albinism {\bf is} prone to sunburn .\\
            \textit{baseline} & Unlike people with high proportion of skin pigment , {\bf people} with albinism {\bf is} prone to sunburn .\\
            \textit{edit selection} & Unlike people with high proportion of skin pigment , {\bf people} with albinism {\bf are} prone to sunburn .\\
            \textit{reference} & Unlike people with high proportion of skin pigment , {\bf people} with albinism {\bf are} prone to sunburn .\\  \hline               

                \multicolumn{2}{|l|}{\textit{Article noun agreement}} \\ \hline
                \textit{source} & As a result , one does not train the necessary parts of the skills required for {\bf a} proper interpersonal {\bf relations} .\\
                \textit{baseline} & As a result , one does not train the necessary parts of the skills required for {\bf a} proper interpersonal {\bf relations} .\\
                \textit{edit selection} & As a result , one does not train the necessary parts of the skills required for proper interpersonal {\bf relations} .\\
                \textit{reference} & As a result , one does not practise the necessary skills required for proper interpersonal {\bf relations} .\\  \hline
                \multicolumn{2}{|l|}{\textit{Verb preposition agreement}} \\ \hline
                \textit{source} & Some people {\bf spend} a lot of time {\bf in} it and forget their real life .\\
                \textit{baseline} & Some people {\bf spend} a lot of time {\bf in} it and forget their real life .\\
                \textit{edit selection} & Some people {\bf spend} a lot of time {\bf on} it and forget their real life .\\
                \textit{reference} & Some people {\bf spend} a lot of time {\bf on} them and forget their real life .\\  \hline
                
            \end{tabular} \par
            \caption[Table]{Example output after edit selection on the 5-best hypotheses compared to the baseline SMT system output. \textit{source} refers to the original erroneous sentence written by an ESL learner. \textit{reference} is the corrected sentence by a human annotator.}
            \label{tab:outputexample}
        \end{center}
    \end{table*}
    
    We observe that both the reranking method and the edit selection  method are able to obtain improvements compared to the baseline. It is worth noting that the reranking method only rescores the n-best hypotheses and selects the hypothesis which attains the highest score. The edit selection method can be seen as a combination of multiple hypotheses, since it is capable of producing a new hypothesis different from any of the top $n$ hypotheses. Our experimental results show that the edit selection method performs better than the reranking method.
    
Our experiments also show that there is no further improvement in reranking top $k$ hypotheses for $k > 10$.  This is because most of the source sentences produce less than 10 hypotheses.  This happens because Moses prunes away a hypothesis during decoding if it gets a low score by the log-linear model. In addition, our analysis shows that lower-ranked hypotheses often contain edits which receive low scores by the edit classifier. Since the average classifier score is used as an additional feature during reranking, lower-ranked hypotheses do not help to improve performance.

    \section{Discussion}
    \label{sub:discuss}
    
    In this section, we discuss the strengths and weaknesses of the SMT approach, showing how our edit selection method helps to improve the system performance. Table~\ref{tab:outputexample} shows a few examples which compare the output of the edit selection method to the output of the baseline system.
    
    Both the subject and verb must agree in number in a sentence. The phrase-based SMT approach to GEC does not take into consideration the interaction between the subject and verb when they are located far from each other. This makes it difficult for the SMT-based baseline to correct such errors. The baseline system also introduces subject-verb agreement errors while attempting to make corrections.  In contrast, our edit classifier considers the interaction between an edit and other phrases and can utilize the n-best hypotheses to produce a better output. In the first example in Table~\ref{tab:outputexample}, the baseline system is not able to correct the verb \emph{is} to \emph{are}. However, the edit selection method considers valid edits from other hypotheses to give a better correction. For example, the nearest noun phrase head to the left of the edit, \emph{people}, helps the classifier gain context knowledge. The edit $is \rightarrow are$ is evaluated as a valid edit.
    
    Similarly, there are other types of errors which are difficult for the SMT approach, but can be overcome by the use of the edit classifier. The second example shows the improper usage of the article \textit{a} for the plural noun \textit{relations}. The interaction between the article and the head noun is not captured by the baseline system because of the distance between the two words in the sentence. The nearest noun phrase head to the right of the edit, which is a feature in our classifier, can capture the dependency between the article and its head noun. The third example also shows a similar weakness of the baseline system when it fails to correct the preposition error.  The nearest verb phrase head to the left of the edit, a feature in the edit classifier, helps to correct the preposition error. 
    
    In addition, we note that edit selection can be seen as a general method to combine multiple correction candidates. Unlike the reranking method, this method can be applied to hypotheses generated by multiple GEC systems. 
    
    \section{Conclusion}
    \label{sec:con}

    In this paper, we have presented our novel approach to improve the accuracy of grammatical error correction, by exploiting the n-best hypotheses of an SMT-based GEC system. Our methods further improve the performance of the SMT approach to GEC, which is one of the dominant approaches for GEC.
    
    We first classify the corrections made by a baseline SMT system as valid or invalid using supervised learning. An edit classifier is built to distinguish between valid versus invalid edits made by the system. We then introduce two novel methods which exploit the list of n-best hypotheses using the edit classifier. The first method is to integrate the classifier's scores into a reranking method. The second method is to select a non-overlapping subset of valid edits which attains the best score. On the CoNLL-2014 test set, both methods give significant improvements compared to a competitive baseline. Our best method achieves an $F_{0.5}$ score of 41.19\% by selecting edits from the list of 5-best hypotheses.
    
    Apart from the hypothesis rank feature in our classifier, all other features of the classifier are independent of the baseline SMT system. This opens up the possibility of using our classifier to combine various correction candidates from multiple GEC systems as well. 
    
    \section*{Acknowledgments}
This research is supported by Singapore Ministry of Education Academic Research Fund Tier 2 grant MOE2013-T2-1-150.  
    
    \bibliographystyle{named}

    \bibliography{ijcai16}
\end{document}